\documentclass[11pt,a4paper]{article}
\usepackage[hyperref]{acl2021}
\usepackage{times}
\usepackage{latexsym}

\usepackage{microtype}
\usepackage{graphicx}
\usepackage{amsmath}
\usepackage{algorithm}
\usepackage{algorithmic}
\usepackage{booktabs}
\usepackage{natbib}

\aclfinalcopy 
\def\aclpaperid{375} 

\title{Relation Classification with Entity Type Restriction}

\author{Shengfei Lyu, \quad Huanhuan Chen\thanks{\, Corresponding author.} \\
  School of Computer Science and Technology \\
  University of Science and Technology of China, Hefei, China  \\
  \texttt{saintfe@mail.ustc.edu.cn, hchen@ustc.edu.cn} }

\date{}

\begin{document}
\maketitle

\begin{abstract}
Relation classification aims to predict a relation between two entities in a sentence. 
The existing methods regard all relations as the candidate relations for the two entities. 
These methods  neglect the restrictions on candidate relations by entity types, 
which leads to some inappropriate relations 
being candidate relations. 
In this paper, we propose a novel paradigm,  RElation Classification with ENtity Type restriction (RECENT), 
which exploits entity types to restrict candidate relations. 
Specially, the mutual restrictions of relations and entity types are formalized and 
introduced into relation classification. Besides, the proposed paradigm, RECENT, 
is model-agnostic. 
Based on two representative models GCN and SpanBERT respectively, 
$\text{RECENT}_{\rm GCN}$ and $\text{RECENT}_{\rm SpanBERT}$ 
 are trained in RECENT\footnote{Our code is available at \url{https://github.com//Saintfe/RECENT}.}.
Experimental results on a standard dataset indicate that 
RECENT improves the performance of GCN and SpanBERT by 6.9 and 4.4 F1 points, respectively. 
Especially, $\text{RECENT}_{\rm SpanBERT}$ achieves a new state-of-the-art on TACRED.

\end{abstract}

\section{Introduction}

Relation classification, a supervised version of relation extraction, aims to predict a relation between two entities in a sentence. 
Relation classification is an important step to construct knowledge bases from a large number of unstructured texts \cite{trisedya-etal-2019-neural}, 
which benefits many natural language processing applications, 
such as natural language generation \cite{kang-hashimoto-2020-improved} and question answering \cite{zhao-2020-conditiion}.

Recently, the majority of methods make use of various neural network architectures to learn a fixed-size representation for a sentence and its entities  
with various language features, such as part of speech (POS), entity types, and dependency trees. 
Dependency trees that are parsed from sentences are exploited by GCN \cite{kipf-2017-gcn} to  model sentences \cite{zhang-etal-2018-graph, guo-etal-2019-attention}. 
As a sequence of words, a sentence is modeled by LSTM \cite{hochreiter1997long} and its entity positions are involved with the attention mechanism \cite{zhang-etal-2017-position}.
More recently, pretrained language models \cite{devlin-etal-2019-bert,baldini-soares-etal-2019-matching,joshi-etal-2020-spanbert} 
achieve good performance in relation classification since they are pretrained on massive corpora.

\begin{figure}[t]
	\centering
	\includegraphics[width=1\columnwidth]{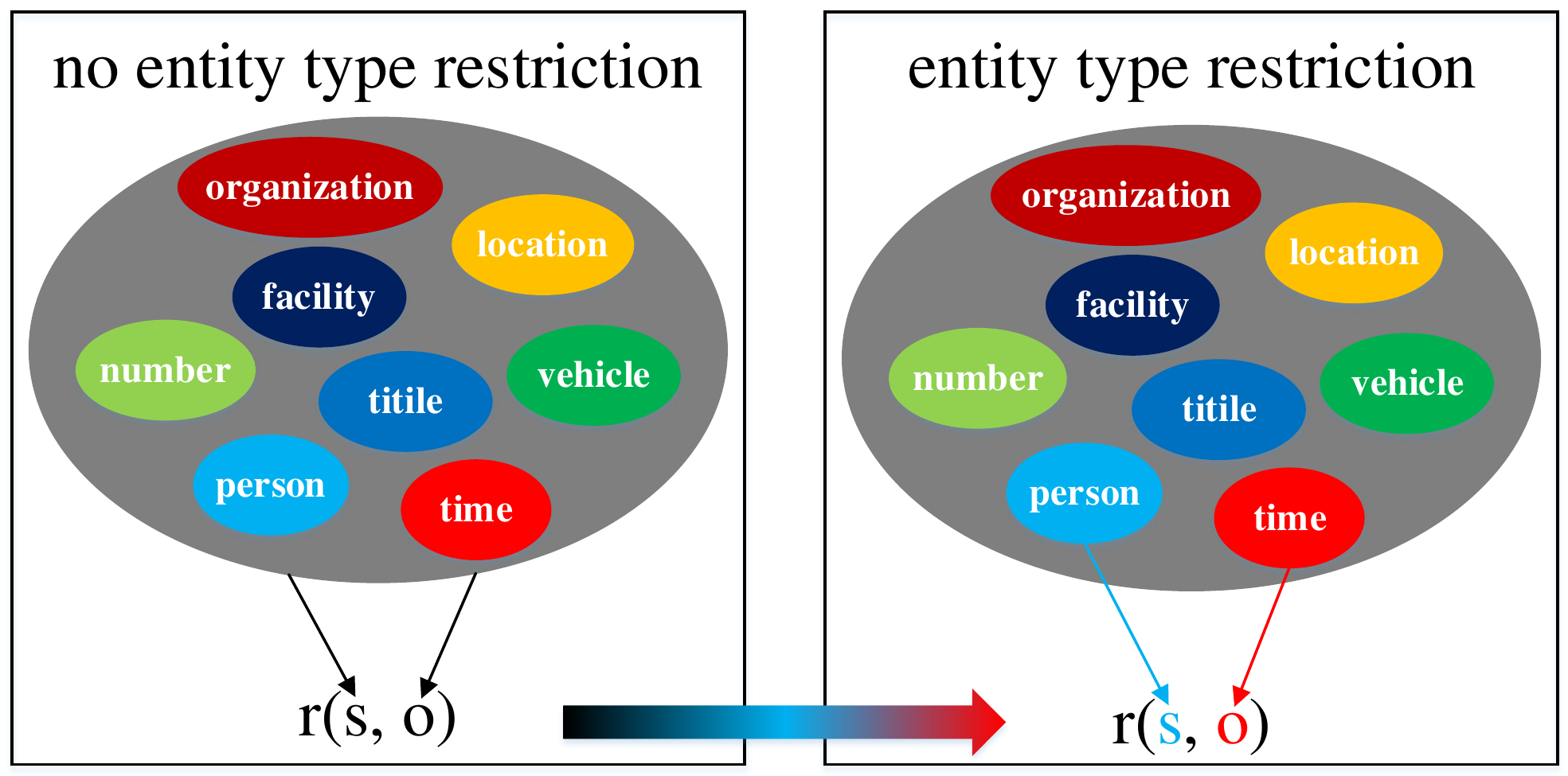}
	\caption{A relation restricts entities  with appropriate types. 
		In the figure,  $r$ is \textit{who-is-born-when}. 
		Different colored ellipses represent entities with different types.
	}
	\label{fig:relation_domain}
\end{figure}

\begin{figure}[t]
	\centering
	\includegraphics[width=1\columnwidth]{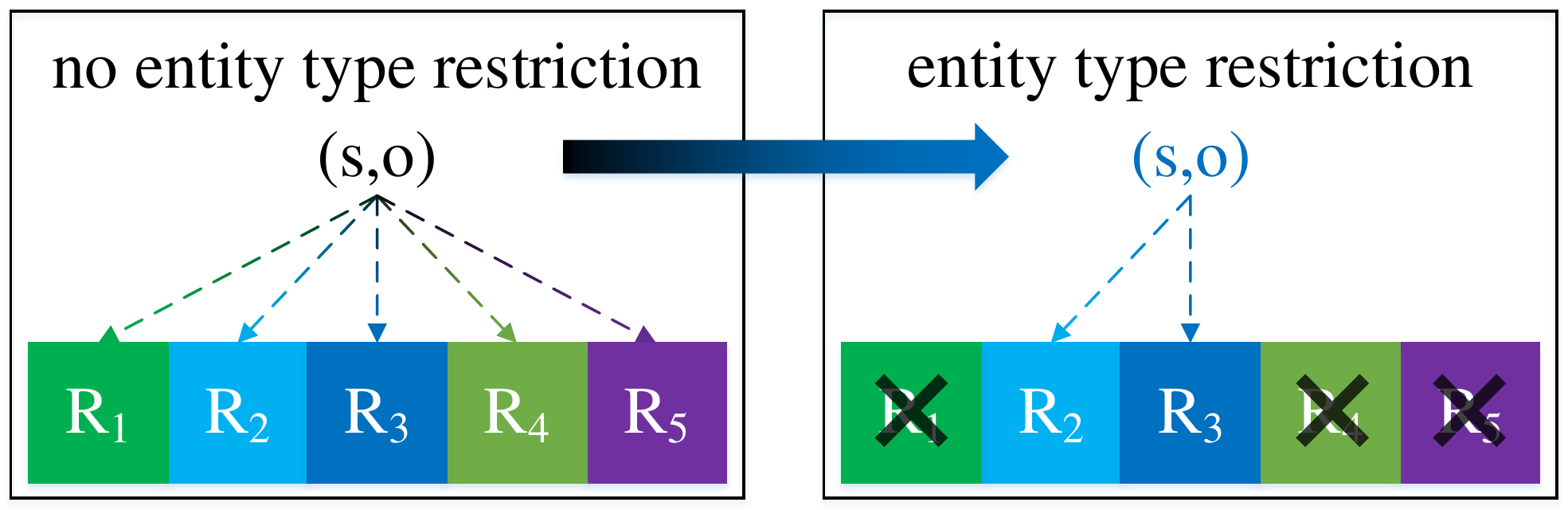}
	\caption{Entity type restriction for relation classification. 
		According to entity type restriction, the number of candidate relations reduces from 5 (left) to 2 (right).}
	\label{fig:entity_type_restriction}
\end{figure}

\begin{figure*}[t]
	\centering
	\includegraphics[width=2\columnwidth]{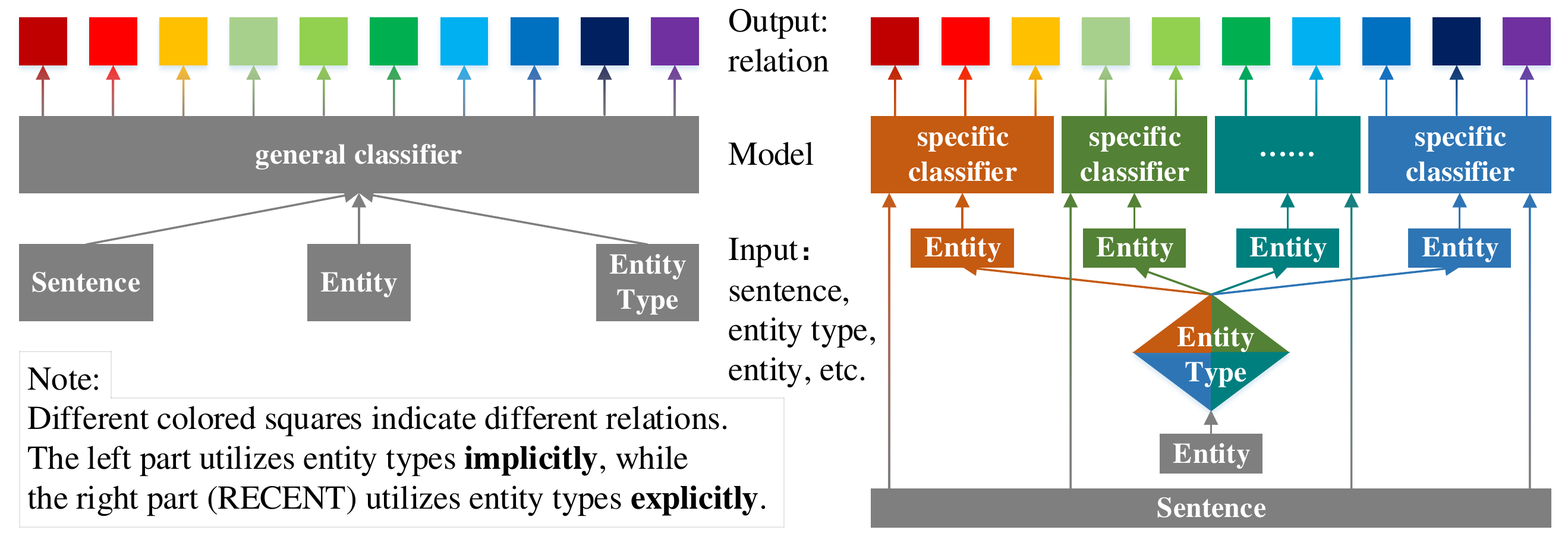}
	\caption{Relation classification with entity type restriction. 
		The left part  does not consider the restriction of entity types on relations and 
		only feeds entity types as features into a general classifier. 
		The right part explicitly utilizes entity types to restrict candidate relations and 
		learns a specific classifier for each pair of entity types.
	}
	\label{fig:rcret}
\end{figure*}

To recap, these methods utilize an encoder architecture \cite{Badrinarayanan-etal-2017-segnet} to obtain a representation for a sentence. 
In other words, they only focus on the modeling of sentences and treat relations as 
labels\footnote{Specifically, these meaningful relations are treated as meaningless numbers, such as 0, 1, 2. } to be classified.
However, in this process, these methods inevitably lose the semantics of relations.  
Take the mutual restrictions between a relation and entity types as an example.
In Figure \ref{fig:relation_domain}, the relation \textit{who-is-born-when} 
restricts its first entity to be a person and the second one to be a time.
Conversely, entity types  can also restrict candidate relations in relation classification.
As illustrated in Figure \ref{fig:entity_type_restriction}, 
some inappropriate relations can be discarded from candidate relations by entity type restriction. 
However, the current methods neglect the restriction of entity types on relations 
so that some inappropriate relations are regarded as candidate relations, 
which further hurts their performance.

To solve the above problem, a novel paradigm, RElation Classification with ENtity Type restriction (RECENT), 
is proposed to exploit entity types to restrict candidate relations.
As the basis of the paradigm, the mutual restrictions of relations and entity types are formalized. 
With the entity type restriction, some inappropriate relations are discarded from the candidate relations of a specific pair of entity types, as illustrated in Figure  \ref{fig:entity_type_restriction}. 
A specific classifier with a specific set of candidate relations is individually learned for each pair of entity types (Figure \ref{fig:rcret}).
Therefore, the proposed paradigm, RECENT, can eliminate the interference from inappropriate candidate relations. 

The contributions are summarized as follows:
\begin{itemize}
	\setlength{\itemsep}{0pt}
	\item The  mutual restrictions of relations and entity types are formalized.
	\item A novel paradigm, RECENT, is proposed to exploit entity types to restrict candidate relations in relation classification.
	\item A new state-of-the-art  is achieved on TACRED.
\end{itemize}


\section{Proposed Paradigm}
Before introducing the proposed paradigm RECENT, 
the mutual restrictions between a relation and a pair of entities are formalized as the basis of RECENT.

\subsection{Relation Function}
When a binary relation is considered as a function, 
this relation has two entities as its two arguments.  
Formally, this relation is formalized as $r(s,o)$, 
where $r$ denotes the \textbf{\underline{r}}elation and $s$, $o$ denote the first (\textbf{\underline{s}}ubject) entity and the second (\textbf{\underline{o}}bject) entity, respectively. 
The range of this relation contains two discrete values \{0, 1\}:
\begin{equation}
r(s,o) = 
\begin{cases}
	1 & r \text{ holds between } s \text{ and } o, \\
	0 & \text{otherwise}.
\end{cases}
\label{eq:relation}
\end{equation}

In a broad sense, the domain of this relation can be any pair of entities. 
However, when a pair of entities with inappropriate types is fed into a specific relation, 
the relation can directly return 0, no need to consider the compositional semantics of the relation and the pair of entities.
For example, a specific relation \textit{who-is-born-when} expects the first argument to be a person and the second one to be a time.
Therefore, (\textit{apple}, \textit{Steven Jobs}) is a pair of inappropriate entities for this relation so that 
\textit{who-is-born-when(apple, Steven Jobs)} returns 0 without considering the compositional semantics, 
since \textit{apple} may refer to either a kind of fruit or a company (not a person) and  \textit{Steven Jobs} may refer to a famous person (not a time). 

Only when a relation receives a pair of appropriate entities whose types match it, 
the combination of the relation and the entities might make sense (i.e., the function defined in Eq. \ref{eq:relation} may return 1).
In this case, it is meaningful to further verify the correctness of the  compositional semantics.
From this perspective, in a narrow sense,  the domain (denoted by $D_r$) of a relation ($r$) is defined as follows:
\begin{equation}
	D_r = \{(s,o)|ts \in S(r) \text{ and } to \in O(r)\},
\end{equation}
where  $ts$ and $to$ denote the \textbf{\underline{t}}ypes of the subject entity ($s$) and the object entity ($o$), respectively.
 $S(r)$ and $O(r)$ are the appropriate types of $r$ on the subject entity ($s$) and the object entity ($o$), respectively.

\subsection{Entity Type Restriction}
In the previous subsection, the narrow domain of a relation restricts entities whose types need to match the relation.
Conversely, given a pair of entities whose types are known, the candidate relations of the entities are also restricted, 
since the match between relations and entity types is mutual.

Formally, given a pair of entities  ($s$, $o$) and their types ($ts$, $to$), 
its candidate relations (denoted by $R_{(ts,to)}$) are restricted into a limited set:
\begin{equation}
	\label{eq:etr}
	\begin{split}
		&R_{(ts,to)} = \{r \in R | {(s,o) \in D_r}\} \\
		&= \{r \in R | ts \in S(r) \text{ and } to \in O(r)\}, 
	\end{split}
\end{equation}
where $R$ denotes all possible relations. 
When the types ($ts$, $to$) of a pair of entities ($s$, $o$) are explicitly utilized to restrict its candidate relations, 
the candidate relations reduce from all possible relations  $R$ into a rather smaller set $R_{(ts,to)}$.

\subsection{Relation Classification}

Unlike traditional methods that classify a sentence and its entities on all candidate relations $R$ (the left part of Figure \ref{fig:rcret}),  
the proposed paradigm, RECENT learns a specific classifier with smaller and more precise candidate relations for each pair of entity types (the right part of Figure \ref{fig:rcret}), 
based on entity type restriction in the previous subsection.

The procedure of RECENT is summarized in Algorithm \ref{alg:recent}.
In the learning phase, all sentences are first grouped by types of their entities (line 1). 
For each group (marked as $g$) with a specific pair of entity types ($ts$,$to$), 
the candidate relations $R_{(ts,to)}$  for the group $g$ are obtained by aggregating the relations in the group $g$ (line 3).
Then, a specific classifier (marked by $f_g$) that maps sentences and their entities in $g$ to $R_{(ts,to)}$, is learned for the group $g$ (line 4).
In the prediction phase, given a new sample ($se$, $s$, $o$, $ts$, $to$), a group (marked as $g^{'}$) is matched by the entity types ($ts$, $to$) (line 6).
Then, the classifier $f_{g^{'}}$ learned on the group $g^{'}$ is utilized to predict a relation according to the input ($se$, $s$, $o$) (line 7).

From the 4th line of Algorithm \ref{alg:recent}, the proposed paradigm RECENT is model-agnostic, 
which means that 
RECENT is theoretically compatible with many relation classification models.

\begin{algorithm}[htb]
	\caption{ RECENT}
	\label{alg:recent}
	\begin{algorithmic}[1] 
		\REQUIRE ~~\\ 
		{\bf Input:} $\mathcal{D} = \{(se_i, s_i,o_i, ts_i,to_i,r_i)|i=1,2,...,N\}$  
		where the subscript $i$ indicates the $i$th sample, $se$ is \textbf{\underline{se}}ntence, $s$ is \textbf{\underline{s}}ubject entity, 
		$o$ is \textbf{\underline{o}}bject entity, $ts$ is \textbf{\underline{t}}ype of \textbf{\underline{s}}ubject entity, $to$ is  \textbf{\underline{t}}ype of \textbf{\underline{o}}bject entity, $r$ is \textbf{\underline{r}}elation. \\
		{\bf Output:} Multiple classifiers.
		\STATE Group sentences by entity types. 
		\FOR{each \textbf{\underline{g}}roup $g$ (enity types ($ts$, $to$)  )}
		\STATE aggregate relations in the group as  candidate relations $R_{(ts,to)}$ defined in Eq. \ref{eq:etr}.
		\STATE learn a classifier (marked as $f_g$) on the group  that maps $\{(se_i, s_i,o_i) \in g\}$ to $R_{(ts,to)}$. 
		\ENDFOR		
		\ENSURE ~~\\ 
		{\bf Input:}  A new sample $\{se, s,o, ts,to\}$,
					each specific classifier for each pair of entity types.	 \\
		{\bf Output:} A relation.
		\STATE match the sample to a group (marked as $g^{'}$) according to the entity types ($ts$, $to$).
		\STATE Use the classifier ($f_{g^{'}}$) learned on the group to map ($se$, $s$, $o$) to a relation. 
		\RETURN the relation. 
	\end{algorithmic}
\end{algorithm}

\section{Experiments}
\subsection{Dataset} 
The proposed paradigm RECENT is evaluated on TACRED\footnote{\url{https://catalog.ldc.upenn.edu/LDC2018T24}} \cite{zhang-etal-2017-position}. 
TACRED contains 41 semantic relations and a special \textit{no\_relation} over 106,264 sentences. 
The subject entities in TACRED are classified  into two types: \textit{PERSON} and \textit{ORGANIZATION} 
while the object entities are categorized into 16 fine-grained types, such as \textit{LOCATION } and \textit{TIME}.
Namely, entity types are known.
By convention, the micro-averaged F1 score (abbreviated as F1) is reported on TACRED.

\subsection{Experimental Setup}

Since \textit{no\_relation}  is a candidate relation of  each pair of entity types in TACRED, 
a binary classifier is first learned to distinguish between 41 semantic relations and \textit{no\_relation}.
In this way, each pair of entity types reduces one candidate relation (i.e. \textit{no\_relation}) in RECENT.
If the binary classifier predicts \textit{no\_relation} for a pair of entities, then the final relation for them is \textit{no\_relation}.
Otherwise, their specific semantic relation is further predicted in  RECENT.

\paragraph{Base Models} The proposed paradigm RECENT is model-agnostic. 
Two representative models 
that are GCN \cite{zhang-etal-2018-graph} and SpanBERT \cite{joshi-etal-2020-spanbert} 
are selected as base models (line 4 in Algorithm \ref{alg:recent}). 
For a fair comparison with a base model, 
all classifiers (including the binary classifier) in RECENT are trained by the base model. 
The corresponding models in the paper are denoted as $\text{RECENT}_{\rm GCN}$ and 	
$\text{RECENT}_{\rm SpanBERT}$.

\paragraph{Hyperparameters}
For $\text{RECENT}_{\rm GCN}$, the path-centric pruning $K$ is set to 1 as GCN \cite{zhang-etal-2018-graph}.
The learning rates for all classifiers in $\text{RECENT}_{\rm GCN}$ are set to 0.3. 
For $\text{RECENT}_{\rm SpanBERT}$, the learning rates for all classifiers are chosen from \{5e-6, 1e-5, 2e-5, 3e-5, 5e-5\} as SpanBERT.

\paragraph{Compared Models} Extensive  models in relation classification are regarded as comparison models. 
They include PA-LSTM \cite{zhang-etal-2017-position}, 
C-GCN  \cite{zhang-etal-2018-graph},
AGGCN  \cite{guo-etal-2019-attention}, 	
C-AGGCN \cite{guo-etal-2019-attention},
MTB  \cite{baldini-soares-etal-2019-matching},
KnowBert   \cite{peters-etal-2019-knowledge}, 
SpanBERT-ALT \cite{lyu-etal-2020-auxiliary},
KEPLER  \cite{wang-etal-2020-kepler}, 
K-Adapter  \cite{wang-etal-2020-k-adapter}, 
and
LUKE  \cite{yamada-etal-2020-luke}. 
To save space, please refer to the original papers of these models for details.

\begin{table}[t]
	\hspace*{-0.2cm} 
	\centering
	\resizebox{8cm}{!}{
		\begin{tabular}{lccc}		
			\toprule
			{Model} & {P} & {R} & {F1}  \\ 
			\midrule
			PA-LSTM $\dagger$ \cite{zhang-etal-2017-position}  & 65.7 & 64.5 & 65.1 \\	
			C-GCN $\dagger$ \cite{zhang-etal-2018-graph}  & 69.9 & 63.3 & 66.4\\
			AGGCN $\dagger$ \cite{guo-etal-2019-attention} & 69.9 & 60.9 & 65.1\\
			C-AGGCN $\dagger$ \cite{guo-etal-2019-attention} & 71.8 & {66.4} & 69.0 \\
			\midrule
			GCN $\dagger$ \cite{zhang-etal-2018-graph} & 69.8 & 59.0 & 64.0 \\
			$\text{RECENT}_{\rm GCN}$ (ours) & 88.3 & 59.3 & {70.9} \\
			\midrule
			\midrule
			SpanBERT-ALT $\dagger$ \cite{lyu-etal-2020-auxiliary} & 69.0 & 73.0 & 70.9 \\
			MTB  $\dagger$ \cite{baldini-soares-etal-2019-matching} & - & - & 71.5 \\			
			KnowBert $\dagger$ \cite{peters-etal-2019-knowledge} & 71.6 & 71.4 & 71.5 \\		
			KEPLER $\dagger$* \cite{wang-etal-2020-kepler} & 71.5 & 72.5 & 72.0 \\			
			K-Adapter $\dagger$* \cite{wang-etal-2020-k-adapter} & 70.14 & 74.04 & 72.04 \\					
			LUKE  $\dagger$ \cite{yamada-etal-2020-luke} & 70.4 & \textbf{75.1} & 72.7 \\
			
			\midrule
			SpanBERT $\dagger$ \cite{joshi-etal-2020-spanbert} & 70.8 & {70.9} & 70.8 \\
			$\text{RECENT}_{\rm SpanBERT}$ (ours) & \textbf{90.9} & 64.2 & \textbf{75.2} \\
			\bottomrule
		\end{tabular}
	}
	\caption{\label{tab:result-rencet} Results on the TACRED dataset. 
		P and R indicate precision and recall, respectively.
		Bold marks the highest values among models.
		$\dagger$ marks results reported in the original papers.
		* marks results from preprint papers.
		}
\end{table}

\subsection{Experimental Results}

The experimental results are presented in Table \ref{tab:result-rencet}. 
$\text{RECENT}_{\rm GCN}$ achieves a significant performance increase on the F1 score above its base model GCN. 
The absolute increase reaches  6.9 from 64.0 to 70.9. 
The main contribution for the F1 increase is the improved precision  that greatly increases from 69.8 to 88.3.
The great increase in precision, which might result from the restriction on candidate relations by entity types in RECENT, indicates the effectiveness of the proposed paradigm RECENT. 
Besides, $\text{RECENT}_{\rm GCN}$ suppresses the compared models that do not include pretrained language models. 

Similarly, $\text{RECENT}_{\rm SpanBERT}$ overtakes its base model SpanBERT by absolute 4.4 points on F1.
The great soar (absolute 20.1 points) on precision contributes the superior F1 of $\text{RECENT}_{\rm SpanBERT}$. 
Unfortunately, 
the decline in recall limits the further improvement of F1.  
This might be due to sample imbalance  of candidate relations, 
which will be further studied in future work. 
On the whole (i.e. F1), $\text{RECENT}_{\rm SpanBERT}$ outperforms all the compared models. 
Especially, $\text{RECENT}_{\rm SpanBERT}$ exceeds the state-of-the-art LUKE 
model\footnote{LUKE achieves the state-of-the-art (72.7) on the published papers. 
	\citet{cohen-etal-2020-relation} report a new state-of-the-art (74.8) in the preprint way. 
	Anyway, $\text{RECENT}_{\rm SpanBERT}$ achieves a new state-of-the-art (75.2). 
} 
by 2.5 F1 points 
and achieves a new state-of-the-art.

\subsection{Error Analysis of GCN}

\begin{table}[t]
	\centering

		\begin{tabular}{lccc|cc}		
			\toprule
			{Model} & {P} & {R} & {F1} & {FP} & {FP(ET)}   \\ 
			\midrule
			GCN  & 68.4 & 60.2 & 64.1 & 1,323 & 144  \\
			\bottomrule
		\end{tabular}
	\caption{\label{tab:result-gcn} Results of our trained GCN on the TACRED dataset. 
		P and R indicate precision and recall, respectively. 
		FP indicates the number of  false positives 
		and FP(ET) indicates the number of false positives that break the entity type restriction. 
	}
\end{table}

This subsection analyzes the influence of a baseline model (i.e. GCN) that neglects the restriction of entity types on relations. 
We retrain a GCN model and the model achieves 68.4 precision, 60.2 recall, and 64.1 F1 (Table \ref{tab:result-gcn}), 
which are similar to the results in its reported paper \cite{zhang-etal-2018-graph}. 
Observing the prediction results of the model, we find that 1) 1,323 examples are false positives in the test set of TACRED, 
2) 144 (about 11\%) false positives among them break the entity type restriction. 
Namely, GCN can make about 89\% of false positives meet the entity type restriction, by implicitly using entity types. 
However, about 11\% of false positives still break the restriction. 
The false positives broken down by relations are counted in Appendix \ref{apx:stat-fp}.
In details, false positives  broken down by relations are weakly negatively correlated with the amount of training data of relations, where the correlation coefficient is -0.39. 
This infers that fewer training examples of relations may lead to more false positives of relations.

\section{Conclusion}

In the paper, a novel paradigm, RECENT, is proposed by entity type restriction.
RECENT reduces  candidate relations for each pair of entity types by 
the mutual restrictions between relations and entity types.
RECENT is model-agnostic. 
$\text{RECENT}_{\rm GCN}$ and $\text{RECENT}_{\rm SpanBERT}$ 
that are based on two representative models GCN and SpanBERT respectively,
outperform their counterparts on the standard dataset TACRED, 
which empirically indicates the effectiveness of the proposed paradigm RECENT.
Especially, $\text{RECENT}_{\rm SpanBERT}$ achieves a new state-of-the-art on TACRED.



\bibliographystyle{acl_natbib}
\bibliography{acl2021}

\begin{thebibliography}{18}
\expandafter\ifx\csname natexlab\endcsname\relax\def\natexlab#1{#1}\fi

\bibitem[{Badrinarayanan et~al.(2017)Badrinarayanan, Kendall, and
  Cipolla}]{Badrinarayanan-etal-2017-segnet}
Vijay Badrinarayanan, Alex Kendall, and Roberto Cipolla. 2017.
\newblock \href {https://doi.org/10.1109/TPAMI.2016.2644615} {{SegNet}: A deep
  convolutional encoder-decoder architecture for image segmentation}.
\newblock \emph{IEEE Transactions on Pattern Analysis and Machine
  Intelligence}, 39(12):2481--2495.

\bibitem[{Baldini~Soares et~al.(2019)Baldini~Soares, FitzGerald, Ling, and
  Kwiatkowski}]{baldini-soares-etal-2019-matching}
Livio Baldini~Soares, Nicholas FitzGerald, Jeffrey Ling, and Tom Kwiatkowski.
  2019.
\newblock \href {https://doi.org/10.18653/v1/P19-1279} {Matching the blanks:
  Distributional similarity for relation learning}.
\newblock In \emph{Proceedings of the 57th Annual Meeting of the Association
  for Computational Linguistics}, pages 2895--2905, Florence, Italy.
  Association for Computational Linguistics.

\bibitem[{Cohen et~al.(2020)Cohen, Rosenman, and
  Goldberg}]{cohen-etal-2020-relation}
Amir D.~N. Cohen, Shachar Rosenman, and Yoav Goldberg. 2020.
\newblock \href {http://arxiv.org/abs/2010.04829} {Relation extraction as
  two-way span-prediction}.
\newblock \emph{CoRR}, abs/2010.04829.

\bibitem[{Devlin et~al.(2019)Devlin, Chang, Lee, and
  Toutanova}]{devlin-etal-2019-bert}
Jacob Devlin, Ming-Wei Chang, Kenton Lee, and Kristina Toutanova. 2019.
\newblock \href {https://www.aclweb.org/anthology/N19-1423} {{BERT}:
  Pre-training of deep bidirectional transformers for language understanding}.
\newblock In \emph{Proceedings of the 2019 Conference of the North {A}merican
  Chapter of the Association for Computational Linguistics: Human Language
  Technologies, Volume 1 (Long and Short Papers)}, pages 4171--4186,
  Minneapolis, Minnesota. Association for Computational Linguistics.

\bibitem[{Guo et~al.(2019)Guo, Zhang, and Lu}]{guo-etal-2019-attention}
Zhijiang Guo, Yan Zhang, and Wei Lu. 2019.
\newblock \href {https://doi.org/10.18653/v1/P19-1024} {Attention guided graph
  convolutional networks for relation extraction}.
\newblock In \emph{Proceedings of the 57th Annual Meeting of the Association
  for Computational Linguistics}, pages 241--251, Florence, Italy. Association
  for Computational Linguistics.

\bibitem[{Hochreiter and Schmidhuber(1997)}]{hochreiter1997long}
Sepp Hochreiter and J{\"u}rgen Schmidhuber. 1997.
\newblock Long short-term memory.
\newblock \emph{Neural computation}, 9(8):1735--1780.

\bibitem[{Joshi et~al.(2020)Joshi, Chen, Liu, Weld, Zettlemoyer, and
  Levy}]{joshi-etal-2020-spanbert}
Mandar Joshi, Danqi Chen, Yinhan Liu, Daniel~S. Weld, Luke Zettlemoyer, and
  Omer Levy. 2020.
\newblock \href {https://doi.org/10.1162/tacl_a_00300} {{S}pan{BERT}: Improving
  pre-training by representing and predicting spans}.
\newblock \emph{Transactions of the Association for Computational Linguistics},
  8:64--77.

\bibitem[{Kang and Hashimoto(2020)}]{kang-hashimoto-2020-improved}
Daniel Kang and Tatsunori Hashimoto. 2020.
\newblock \href {https://doi.org/10.18653/v1/2020.acl-main.66} {Improved
  natural language generation via loss truncation}.
\newblock In \emph{Proceedings of the 58th Annual Meeting of the Association
  for Computational Linguistics}, pages 718--731, Online. Association for
  Computational Linguistics.

\bibitem[{Kipf and Welling(2017)}]{kipf-2017-gcn}
Thomas~N. Kipf and Max Welling. 2017.
\newblock Semi-supervised classification with graph convolutional networks.
\newblock In \emph{Proceedings of the 5th International Conference on Learning
  Representations, Toulon, France}.

\bibitem[{{Lyu} et~al.(2020){Lyu}, {Cheng}, {Wu}, {Cui}, {Chen}, and
  {Miao}}]{lyu-etal-2020-auxiliary}
S.~{Lyu}, J.~{Cheng}, X.~{Wu}, L.~{Cui}, H.~{Chen}, and C.~{Miao}. 2020.
\newblock \href {https://doi.org/10.1109/TETCI.2020.3040444} {Auxiliary
  learning for relation extraction}.
\newblock \emph{IEEE Transactions on Emerging Topics in Computational
  Intelligence}, pages 1--10.

\bibitem[{Peters et~al.(2019)Peters, Neumann, Logan, Schwartz, Joshi, Singh,
  and Smith}]{peters-etal-2019-knowledge}
Matthew~E. Peters, Mark Neumann, Robert Logan, Roy Schwartz, Vidur Joshi,
  Sameer Singh, and Noah~A. Smith. 2019.
\newblock \href {https://doi.org/10.18653/v1/D19-1005} {Knowledge enhanced
  contextual word representations}.
\newblock In \emph{Proceedings of the 2019 Conference on Empirical Methods in
  Natural Language Processing and the 9th International Joint Conference on
  Natural Language Processing (EMNLP-IJCNLP)}, pages 43--54, Hong Kong, China.
  Association for Computational Linguistics.

\bibitem[{Trisedya et~al.(2019)Trisedya, Weikum, Qi, and
  Zhang}]{trisedya-etal-2019-neural}
Bayu~Distiawan Trisedya, Gerhard Weikum, Jianzhong Qi, and Rui Zhang. 2019.
\newblock \href {https://doi.org/10.18653/v1/P19-1023} {Neural relation
  extraction for knowledge base enrichment}.
\newblock In \emph{Proceedings of the 57th Annual Meeting of the Association
  for Computational Linguistics}, pages 229--240, Florence, Italy. Association
  for Computational Linguistics.

\bibitem[{Wang et~al.(2020{\natexlab{a}})Wang, Tang, Duan, Wei, Huang, Ji, Cao,
  Jiang, and Zhou}]{wang-etal-2020-k-adapter}
Ruize Wang, Duyu Tang, Nan Duan, Zhongyu Wei, Xuanjing Huang, Jianshu Ji,
  Guihong Cao, Daxin Jiang, and Ming Zhou. 2020{\natexlab{a}}.
\newblock \href {http://arxiv.org/abs/2002.01808} {K-{A}dapter: Infusing
  knowledge into pre-trained models with adapters}.
\newblock \emph{CoRR}, abs/2002.01808v5.

\bibitem[{Wang et~al.(2020{\natexlab{b}})Wang, Gao, Zhu, Liu, Li, and
  Tang}]{wang-etal-2020-kepler}
Xiaozhi Wang, Tianyu Gao, Zhaocheng Zhu, Zhiyuan Liu, Juanzi Li, and Jian Tang.
  2020{\natexlab{b}}.
\newblock \href {http://arxiv.org/abs/1911.06136} {{KEPLER:} {A} unified model
  for knowledge embedding and pre-trained language representation}.
\newblock \emph{CoRR}, abs/1911.06136v3.

\bibitem[{Yamada et~al.(2020)Yamada, Asai, Shindo, Takeda, and
  Matsumoto}]{yamada-etal-2020-luke}
Ikuya Yamada, Akari Asai, Hiroyuki Shindo, Hideaki Takeda, and Yuji Matsumoto.
  2020.
\newblock \href {https://doi.org/10.18653/v1/2020.emnlp-main.523} {{LUKE}: Deep
  contextualized entity representations with entity-aware self-attention}.
\newblock In \emph{Proceedings of the 2020 Conference on Empirical Methods in
  Natural Language Processing (EMNLP)}, pages 6442--6454, Online. Association
  for Computational Linguistics.

\bibitem[{Zhang et~al.(2018)Zhang, Qi, and Manning}]{zhang-etal-2018-graph}
Yuhao Zhang, Peng Qi, and Christopher~D. Manning. 2018.
\newblock \href {https://doi.org/10.18653/v1/D18-1244} {Graph convolution over
  pruned dependency trees improves relation extraction}.
\newblock In \emph{Proceedings of the 2018 Conference on Empirical Methods in
  Natural Language Processing}, pages 2205--2215, Brussels, Belgium.
  Association for Computational Linguistics.

\bibitem[{Zhang et~al.(2017)Zhang, Zhong, Chen, Angeli, and
  Manning}]{zhang-etal-2017-position}
Yuhao Zhang, Victor Zhong, Danqi Chen, Gabor Angeli, and Christopher~D.
  Manning. 2017.
\newblock \href {https://doi.org/10.18653/v1/D17-1004} {Position-aware
  attention and supervised data improve slot filling}.
\newblock In \emph{Proceedings of the 2017 Conference on Empirical Methods in
  Natural Language Processing}, pages 35--45, Copenhagen, Denmark. Association
  for Computational Linguistics.

\bibitem[{Zhao et~al.(2020)Zhao, Xiao, Zhong, Yao, and
  Chen}]{zhao-2020-conditiion}
Xinyan Zhao, Feng Xiao, Haoming Zhong, Jun Yao, and Huanhuan Chen. 2020.
\newblock \href {https://doi.org/10.1145/3366423.3380301} {Condition aware and
  revise transformer for question answering}.
\newblock In \emph{Proceedings of The Web Conference}, page 2377–2387.

\end{thebibliography}

\appendix 
\section{The Statistics of False Positives}
\label{apx:stat-fp} 

\begin{table}[t]
	\centering
	\begin{tabular}{lc}
		\toprule
		{Relation} & {False Positives}   \\ 
		\midrule
		org:alternate\_names                  & 53 \\
		org:city\_of\_headquarters            & 30 \\
		org:country\_of\_headquarters         & 70 \\
		org:dissolved                         & 2  \\
		org:founded                           & 9  \\
		org:founded\_by                       & 46 \\
		org:member\_of                        & 18 \\
		org:members                           & 31 \\
		org:number\_of\_employees/members     & 9  \\
		org:parents                           & 62 \\
		org:political/religious\_affiliation  & 4  \\
		org:shareholders                      & 10 \\
		org:stateorprovince\_of\_headquarters & 16 \\
		org:subsidiaries                      & 37 \\
		org:top\_members/employees            & 61 \\
		org:website                           & 1  \\
		per:age                               & 13 \\
		per:alternate\_names                  & 11 \\
		per:cause\_of\_death                  & 37 \\
		per:charges                           & 35 \\
		per:children                          & 31 \\
		per:cities\_of\_residence             & 90 \\
		per:city\_of\_birth                   & 3  \\
		per:city\_of\_death                   & 19 \\
		per:countries\_of\_residence          & 93 \\
		per:country\_of\_birth                & 5  \\
		per:country\_of\_death                & 9  \\
		per:date\_of\_birth                   & 3  \\
		per:date\_of\_death                   & 36 \\
		per:employee\_of                      & 98 \\
		per:origin                            & 46 \\
		per:other\_family                     & 60 \\
		per:parents                           & 49 \\
		per:religion                          & 16 \\
		per:schools\_attended                 & 13 \\
		per:siblings                          & 25 \\
		per:spouse                            & 24 \\
		per:stateorprovince\_of\_birth        & 4  \\
		per:stateorprovince\_of\_death        & 10 \\
		per:stateorprovinces\_of\_residence   & 40 \\
		per:title                             & 94 \\
		\bottomrule
	\end{tabular}
	\caption{\label{tab:stat-fp-tacred} False positives broken down by relations of our trained GCN on the TACRED dataset. }
\end{table}
Table \ref{tab:stat-fp-tacred} presents  false positives broken down by relations of our trained GCN on the TACRED dataset. 
In details, false positives  broken down by relations are weakly negatively correlated with the amount of training data of relations, where the correlation coefficient is -0.39. 

\end{document}